%% file: main_ijcai.tex
\documentclass{article}
\usepackage{LearnSim/lib/ijcai21}

\input{LearnSim/redefine_ijcai}
\input{LearnSim/title}

\begin{document}
\maketitle
\input{LearnSim/abstract}
\input{LearnSim/introduction}
\input{LearnSim/related_work}
\input{LearnSim/problem_definition}
\input{LearnSim/method}

\input{LearnSim/experiment}

\input{LearnSim/discussion}

\input{LearnSim/conclusion}
\input{LearnSim/acknowledment}

\clearpage
\bibliographystyle{LearnSim/lib/ijcai21.bst}
  \bibliography{LearnSim/bibfile.bib}

\clearpage
\input{LearnSim/appendix}

\end{document}

%% file: LearnSim/redefine_ijcai.tex
\usepackage{times}

\usepackage{soul}
\usepackage{url}
\usepackage[hidelinks]{hyperref}
\usepackage[utf8]{inputenc}
\usepackage[small]{caption}
\usepackage{graphicx}
\usepackage{amsmath}
\usepackage{booktabs}
\usepackage{textcomp}
\usepackage{xspace}
\usepackage{mathrsfs}
\usepackage{enumitem}
\usepackage{url}
\usepackage{hyperref}
\usepackage{appendix}

\usepackage{xcolor}
\usepackage{todonotes}

\usepackage[linesnumbered,ruled,vlined]{algorithm2e}
\usepackage{multirow}
\usepackage{tabularx}
\usepackage{graphicx}
\usepackage{float}
\usepackage{subfigure}
\usepackage{array}
\usepackage{amsfonts}
\usepackage{epsfig}
\usepackage{graphicx}
\usepackage{amsmath}
\usepackage{amssymb}
\usepackage{booktabs}
\usepackage{color}
\usepackage{flushend}

\newcommand{\PreserveBackslash}[1]{\let\temp=\\#1\let\\=\temp}
\newcolumntype{C}[1]{>{\PreserveBackslash\centering}p{#1}}
\newcolumntype{R}[1]{>{\PreserveBackslash\raggedleft}p{#1}}
\newcolumntype{L}[1]{>{\PreserveBackslash\raggedright}p{#1}}

\newcommand{\hz}{HZ\xspace}

\newcommand{\gd}{GD\xspace}

\newcommand{\bc}{BC\xspace}
\newcommand{\deepirl}{DeepIRL\xspace}
\newcommand{\gail}{MA-GAIL\xspace}
\newcommand{\ours}{PS-AIRL\xspace}
\newcommand{\cfmrs}{CFM-RS\xspace}
\newcommand{\cfmts}{CFM-TS\xspace}

\newcommand{\nop}[1]{}

\newtheorem{problem}{Problem}

\setlist[itemize]{leftmargin=*}

%% file: LearnSim/title.tex
\title{Objective-aware Traffic Simulation via Inverse Reinforcement Learning}

\author{
Guanjie Zheng$^{1,4 *}$ \and
Hanyang Liu$^{2}$\thanks{Equal contribution. Part of the work was done when Guanjie Zheng and Hanyang Liu were interns at Tianrang Inc. and when Guanjie Zheng was completing his PhD at the Pennsylvania State University.} \and
Kai Xu$^3$ \and
Zhenhui Li$^4$ \\
\affiliations
$^1$Shanghai Jiao Tong University, $^2$Washington University in St. Louis \\
$^3$Tianrang Inc., $^4$The Pennsylvania State University \\
\emails
gjzheng@sjtu.edu.cn, hanyang.liu@wustl.edu, kai.xu@tianrang-inc.com, jessieli@ist.psu.edu
}

%% file: LearnSim/abstract.tex
%
\begin{abstract} 
Traffic simulators act as an essential component in the operating and planning of transportation systems. Conventional traffic simulators usually employ a calibrated physical car-following model to describe vehicles' behaviors and their interactions with traffic environment. However, there is no universal physical model that can accurately predict the pattern of vehicle's behaviors in different situations. A fixed physical model tends to be less effective in a complicated environment given the non-stationary nature of traffic dynamics. In this paper, we formulate traffic simulation as an inverse reinforcement learning problem, and propose a parameter sharing adversarial inverse reinforcement learning model for dynamics-robust simulation learning. Our proposed model is able to imitate a vehicle's trajectories in the real world while simultaneously recovering the reward function that reveals the vehicle's true objective which is invariant to different dynamics. Extensive experiments on synthetic and real-world datasets show the superior performance of our approach compared to state-of-the-art methods and its robustness to variant dynamics of traffic.
\end{abstract}

%% file: LearnSim/introduction.tex
\section{Introduction}
\label{sec:introduction}

Traffic simulation has long been one of the significant topics in transportation. Microscopic traffic simulation plays an important role in planing, designing and operating of transportation systems. For instance, an elaborately designed traffic simulator allows the city operators and planers to test policies of urban road planning, traffic restrictions, and the optimization of traffic congestion, by accurately deducing possible effects of the applied policies to the urban traffic environment ~\cite{toledo2003calibration}. The prior work \cite{wei2018intellilight} employs a traffic simulator to train and test policies of intelligent traffic signal control, as it can generate a large number of simulated data for the training of the signal controller.

\begin{figure}[t]
\centering
 \includegraphics[width=0.7\linewidth]{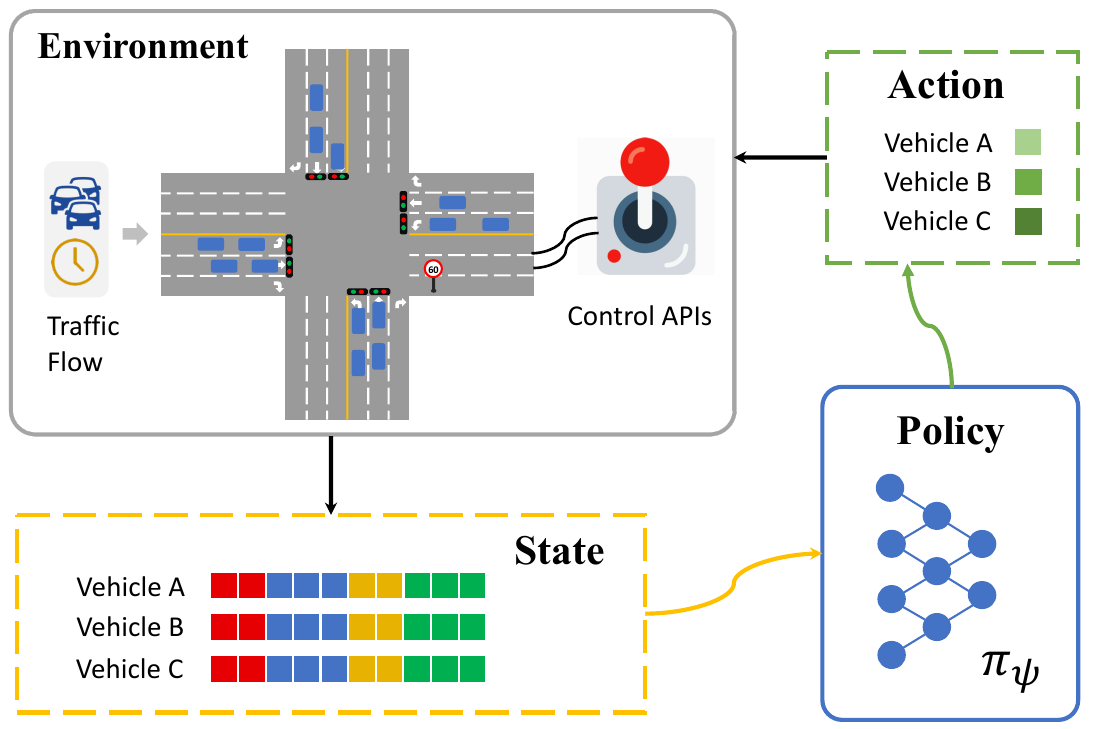} 
\caption{Interactions between the simulation environment and policy (e.g., car-following model). }
\label{fig:environment}
\end{figure}

Current transportation approaches used in the state-of-the-art traffic simulators such as~\cite{yu2017calibration,osorio2019efficient} employ several physical and empirical equations to describe the kinetic movement of individual vehicles, referred to as the car-following model (CFM). 
The parameters of CFMs such as max acceleration and driver reaction time must be carefully calibrated using traffic data. A calibrated CFM can be exploited as a policy providing the vehicle optimal actions given the state of the environment, as shown in Figure \ref{fig:environment}. The optimality of this policy is enabled by the parameter calibration that obtains a close match between the observed and simulated traffic measurements.

An effective traffic simulator should produce accurate simulations despite variant dynamics in different traffic environment. This can be factorized into two specific objectives. The {\em first objective} is to accurately imitate expert vehicle behaviors given a certain environment with stationary dynamics. 
The movement of real-world vehicles depends on many factors including speed, distance to neighbors, road networks, traffic lights, and also driver's psychological factors.
A CFM usually aims to imitate the car-following behavior by applying physical laws and human knowledge in the prediction of vehicle movement. Faced with sophisticated environment, models with emphasis on fitting different factors are continuously added to the CFM family. For example, Krauss model~\cite{krauss1998microscopic} focuses on safety distance and speed, while Fritzsche model sets thresholds for vehicle's movement according to driver's psychological tendency. However, there is no universal model that can fully uncover the truth of vehicle-behavior patterns under comprehensive situations. Relying on inaccurate prior knowledge, despite calibrated, CFMs often fail to exhibit realistic simulations.

The {\em second objective} is to make the model robust to variant dynamics in different traffic environment.
However, it is challenging due to the non-stationary nature of real-world traffic dynamics. For instance, weather shifts and variances of road conditions may change a vehicle's mechanical property and friction coefficient against the road surface, and eventually lead to variances of its acceleration and braking performance. In a real-world scenario, a vehicle would accordingly adjust its driving policy and behave differently (e.g., use different acceleration or speed given the same observation) under these dynamics changes. However, given a fixed policy (i.e., CFM), current simulators in general, fail to adapt policies to different dynamics. To simulate a different traffic environment with significantly different dynamics, the CFM must be re-calibrated using new trajectory data with respect to that environment, which is inefficient and sometimes unpractical. This makes these simulation models less effective when generalized to changed dynamics.

We aim to achieve both objectives.
For the first objective, a natural consideration is to learn patterns of vehicles' behaviors directly from real-world observations, instead of relying on sometimes unreliable prior knowledge. Recently, imitation learning (IL) has shown promise for learning from demonstrations~\cite{ho2016generative,fu2017learning}. 
However, direct IL methods such as behavioral cloning~\cite{michie1990cognitive} that aim to directly extract an expert policy from data, would still fail in the second objective, as the learned policy may lose optimality when traffic dynamics change.
A variant of IL, inverse reinforcement learning (IRL), different from direct IL, not only learns an expert's policy, but also infers the reward function (i.e., cost function or objective) from demonstrations. The learned reward function can help explain an expert's behavior and give the agent an objective to take actions imitating the expert, which enables IRL to be more interpretable than direct IL. 
To this end, we propose to use an IRL-based method that build off the adversarial IRL (AIRL)~\cite{fu2017learning}, to train traffic simulating agents that both generate accurate trajectories while simultaneously recovering the invariant underlying reward of the agents, which facilitates the robustness to variant dynamics.

With disentangled reward of an agent recovered from real-world demonstrations, we can infer the vehicle's true objective. Similar to the human driver's intention (e.g., drive efficiently under safe conditions), the estimated objective can be invariant to changing dynamics (e.g., maximum acceleration or deceleration). 
\nop{We can further use the recovered reward function to optimize new policies optimal for changed dynamics.}
Considering the intricate real-world traffic with multiple vehicles interacting with each other, we extend AIRL to the multi-agent context of traffic simulation. We incorporate a scalable decentralized parameter sharing mechanism in~\cite{gupta2017cooperative} with AIRL, yielding a new algorithm called Parameter Sharing AIRL (PS-AIRL), as a dynamics-robust traffic simulation model. 
In addition, we propose an online updating procedure, using the learned reward to optimize new policies to adapt to different dynamics in the complex environment, without any new trajectory data.
Specifically, our contributions are as threefold:
\begin{itemize}
    \vspace{-0.3em}
    \item We propose an IRL-based model that is able to infer real-world vehicle's true objective. It enables us to optimize policies adaptive and robust to different traffic dynamics. 
    \vspace{-0.3em}
    \item We extend the proposed model with the parameter sharing mechanism to a multi-agent context, enabling our model good scalable capacity in large traffic. 
    \vspace{-0.3em}
    \item Extensive experiments on both synthetic and real-world datasets show the superior performance in trajectory simulation, reward recovery and dynamics-robustness of \ours over state-of-the-art methods.
\end{itemize}

%% file: LearnSim/related_work.tex
\section{Related Work}
\label{sec:related-work}

\textbf{Conventional CFM Calibration} 
Traditional traffic simulation methods such as SUMO~\cite{SUMO2012}, AIMSUN~\cite{barcelo2005dynamic} and MITSIM~\cite{yang1996microscopic}, are based on calibrated CFM models, which are used to simulate the interactions between vehicles. In general, CFM considers a set of features to determine the speed or acceleration, such as safety distance, and velocity difference between two adjacent vehicles in the same lane. 

\textbf{Learning to Simulate} Recently, several approaches proposed to apply reinforcement learning in simulation learning problems such as autonomous driving~\cite{wang2018deep}. For these works, the reward design is necessary but complicated because it is difficult to mathematically interpret the vehicle's true objective. The prior work \cite{bhattacharyya2018multi} proposed to frame the driving simulation problem in multi-agent imitation learning, and followed the framework of~\cite{ho2016generative} to learn optimal driving policy from demonstrations, which has a similar formulation to our paper. But different from~\cite{bhattacharyya2018multi}, our traffic simulation problem has totally different concentrations. Our work mainly focus on the fidelity of simulated traffic flow and car-following behavior, as well as correct reactions to traffic light switchover, while \cite{bhattacharyya2018multi} ignores traffic signals, and underlines safe driving when interacting with neighboring vehicles. The work \cite{zheng2020learning} uses a similar framework for traffic simulation but is unable to achieve dynamic-robustness like our proposed model. 

\begin{figure*}[t!]
\centering
 \includegraphics[width=0.9\textwidth]{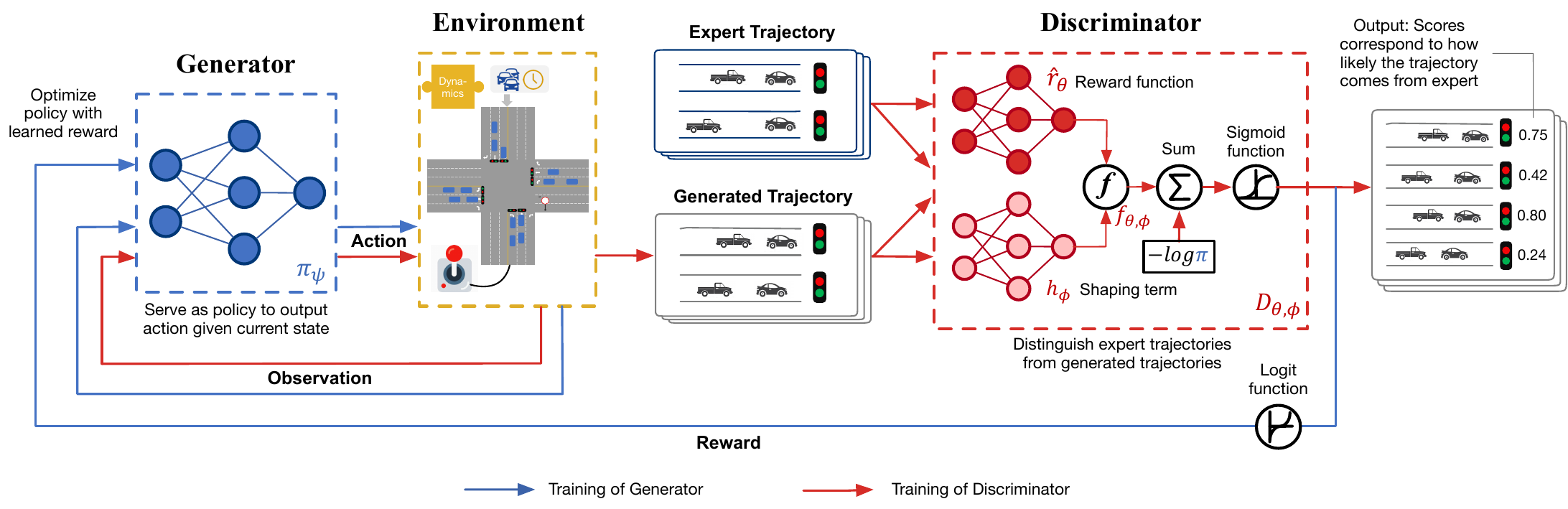}
 \vspace{-3mm}
\caption{Overview of PS-AIRL. The discriminator and the generator are trained alternatively. When training the discriminator, the generator serves as a fixed policy that rollouts trajectories. The discriminator takes both generated and expert trajectories as input and distinguishes between them. When training the generator, the learned reward function is used as the cost function to update the optimal policy.}
\label{fig:pipeline}
\end{figure*}

%% file: LearnSim/problem_definition.tex
\section{Problem Definition}
\label{sec:problem-definition}

We frame traffic simulation as a multi-agent control problem,
regarding the complicated interaction between vehicles in traffic.
Formally, we formulate our model in a decentralized partially observable Markov decision process (DPOMDP) defined by the tuple $(\mathcal{M}, \{\mathcal{S}_m\}, \{\mathcal{A}_m\}, \mathcal{T}, r, \gamma, \rho_0)$. Here $\mathcal{M}$ denotes a finite set of agents, $\{\mathcal{S}_m\}$ and $\{\mathcal{A}_m\}$ the sets of states and actions for each agent $m$, respectively. We have $\mathcal{S}_m \in \mathbb{R}^S$ and $\mathcal{A}_m \in \mathbb{R}^A$ for all agents $m\in\mathcal{M}$. $\rho_0$ denotes the initial state distribution. $r(s,a)$ is the reward function, and $\gamma$ denotes the long-term reward discount factor. We assume for a set of given expert demonstrations, the environment dynamics remain unchanged. The deterministic state transition function is defined by $\mathcal{T} (s'|s,a)$.
We formulate the microscopic traffic simulation problem in a inverse reinforcement learning (IRL) fashion. Given trajectories of expert vehicles, our goal is to learn the reward function of the vehicle agent. Formally, the problem is defined as follows:

\begin{problem}
Given a set of expert trajectories as demonstrations $\mathcal{D} = \{\tau_1, \tau_2, ..., \tau_M\}$ generated by an optimal policy $\pi^*(a|s)$ where $\tau=\{s_1,a_1,...,s_T,a_T\}$, the goal is to recover the reward function $r_{\theta}(s,a)$, so that the log likelihood of the expert trajectories are maximized, i.e.,
\begin{equation}
\max_\theta E_{\tau \sim \mathcal{D}} [\log p_\theta(\tau)]
\label{eq:log-like}
\end{equation}
Here, $p_{\theta}(\tau) \propto \exp \{\sum_{t=0}^T\gamma^{t} r_{\theta}\left(s_{t}, a_{t}\right)\}$ is the distribution of trajectories parameterized by $\theta$ in the reward $r_\theta(s,a)$.
\end{problem}

%% file: LearnSim/method.tex
\section{Method}
\label{sec:method}

\subsection{Adversarial Inverse Reinforcement Learning}
\label{sec:method-airl}

Along the line of IRL, ~\cite{finn2016guided} proposed to treat the optimization problem in Eq.~\eqref{eq:log-like} as a GAN-like~\cite{goodfellow2014generative} optimization problem. 
This adversarial training framework alternates between training the discriminator to classify each expert trajectory and updating the policy to \textit{adversarially} confuse the discriminator, and optimize the loss: 
\begin{equation}
    \min_\theta \max_\psi {\mathbb{E}}_{\pi_\psi}\left[\log \left(D_{\theta}(s, a)\right)\right]+{\mathbb{E}}_{\pi_E}\left[ \log \left(1-D_{\theta}(s, a)\right)\right]
    \label{eq:gail-discrim}
\end{equation}
where $D_\theta$ is the discriminator parameterized by $\theta$, $\pi_E$ is the optimal policy for expert demonstration data $\mathcal{D}$, and $\pi_\psi$ is the learned policy parameterized by $\psi$ in the generator. 
Specifically, the discriminator scores each sample with its likelihood of being from the expert, and use this score as the learning cost to train a policy that generates fake samples. Thus, the objective of the generator is to generate samples that confuse the discriminator, where $\log D_\theta(s,a)$ is used as a surrogate reward to help guide the generated policy into a region close to the expert policy. Its value grows larger as samples generated from $\pi_\psi$ look similar to those in expert data.

We use the same strategy to train a discriminator-generator network. However, one limitation of \cite{finn2016guided} is that, the use of full trajectories usually cause high variance. Instead, we follow \cite{finn2016connection} to use state-action pairs as input with a more straightforward conversion for the discriminator:
\begin{equation}
\label{eq:discriminator-design}
    D(s, a) = \frac{\exp\{f(s, a)\}}{\exp\{f(s, a)\} + \pi(a|s)}
\end{equation}
where $f$ is a learned function, and policy $\pi$ is trained to maximize reward $r(s, a) = \log(1-D(s,a))-\log D(s,a)$. In each iteration, the extracted reward serves as the learning cost for training the generator to update the policy. Updating the discriminator is equivalent to updating the reward function, and in turn updating the policy can be viewed as improving the sampling distribution used to estimate the discriminator. 
The structure of the discriminator in Eq. \eqref{eq:discriminator-design} can be viewed as a sigmoid function with input $f(s,a)-\log\pi$, as shown in Fig. \ref{fig:pipeline}. Similar to GAN, the discriminator is trained to reach the optimality when the expected $D(s,a)$ over all state-action pairs is $1/2$. At the optimality, $f^*(s,a)=\log\pi^*(a|s)$ and the reward can be extracted from the optimal discriminator by $r(s, a) = f(s,a) + Const$.

\subsection{Dynamics-robust Reward Learning}
We seek to learn the optimal policy for the vehicle agent from real-world demonstrations. So the learned policy should be close to, in the real world, the driver's policy to control the movement of vehicles given the state of traffic environment.
Despite the non-stationary nature of traffic dynamics, the vehicle's objective during driving should constantly remain the same. This means that although the optimal policy under different system dynamics may vary, the vehicle constantly has the same reward function. Therefore, it is highly appealing to learn an underlying dynamic-robust reward function beneath the expert policy. By defining the traffic simulation learning as an IRL problem, we adapt the adversarial IRL (AIRL) as in Section \ref{sec:method-airl} on the traffic simulation problem and propose a novel simulation model, which serves to not only imitate the expert trajectories (i.e., driving behaviors), but simultaneously recover the underlying reward function.

We aim to utilize the reward-invariant characteristics of AIRL to train an agent that simulates a vehicle's behavior in traffic while being robust to dynamics with changing traffic environment. The reward ambiguity is an essential issue in IRL approaches. \cite{ng1999policy} proposes a class of reward transformations that preserve the optimal policy, for any function $\Phi:\mathcal{S}\longrightarrow\mathbb{R}$, we have
\begin{equation}
    \hat{r}(s,a,s')\in\{\hat{r} | \hat{r}=r(s,a,s')+\gamma\Phi(s')-\Phi(s)\}
\end{equation}
If the true reward is solely a function of state, we are able to extract a reward that is fully disentangled from dynamics (i.e., invariant to changing dynamics) and recover the ground truth reward up to a constant~\cite{fu2017learning}. 

To learn this \textit{dynamic-robust reward} and ensure it remains in the set of rewards that correspond to the same optimal policy, AIRL follows the above reward transformation, and replaces the learned function $f(s,a)$ in Eq. (\ref{eq:discriminator-design}) with
\begin{equation}
\label{eq:learned-func}
f_{\theta,\phi}(s,a,s')=\hat{r}_\theta(s) + \gamma h_\phi(s')-h_\phi(s)
\end{equation}
where $\hat{r}_\theta(s)$ is the state-only reward approximator, $h_\phi$ is a shaping term, and $s'=\pi(s,a)$ is the next state. The transformation in Eq. \eqref{eq:learned-func} ensures that the learned function $f_{\theta,\phi}$ corresponds to the same optimal policy as the reward approximator $\hat{r}_\theta$ does. Correspondingly, $D_\theta(s,a)$ in Eq. \eqref{eq:discriminator-design} becomes
\begin{equation}
    D_{\theta,\phi}(s,a,s')=\frac{\exp\{f_{\theta\phi}(s, a, s')\}}{\exp\{f_{\theta\phi}(s, a, s')\} + \pi(a|s)}
\end{equation}
As a result, the discriminator can be represented by the combination of the two functions $\hat{r}_\theta$ and $h_\phi$ as shown in Fig. \ref{fig:pipeline}.

\subsection{Parameter-sharing Agents}
\label{sec:method-multi-agent}

\begin{algorithm}[tb]
\DontPrintSemicolon 
\caption{Training procedure of PS-AIRL}

\KwIn{Expert trajectories $\tau_E \sim \pi_E$}
\KwOut{Policy $\pi_\psi$, reward $r_{\theta, \phi}$}

Initialize policy parameters $\psi$, and discriminator parameters $\theta$ and $\phi$\;
\For{k $\longleftarrow$ 0, 1, \dots}
{
    Rollout trajectories for all $M$ agents $\vec{\tau}=\{\tau_1, \tau_2, ..., \tau_M\} \sim \pi_{\psi_k}$ \;
    
    Update $\theta$, $\phi$ in discriminator given $\vec{\tau}$ via minimizing ${\mathbb{E}}_{\pi_\psi}\!\!\left[\log \left(D_{\theta,\phi}(s,a,s')\right)\right] \!+\! {\mathbb{E}}_{\pi_E}\!\!\left[ \log \left(1 \!\!- \!\!D_{\theta,\phi}(s,a,s')\right)\right]$\;
    
    Update reward using discriminator output 
    $r_{\theta, \phi}\longleftarrow\log D_{\theta,\phi}- \log (1 - D_{\theta,\phi})$ \;
    
    Update policy $\pi_\psi$ with a TRPO step by solving the optimization in Eq. \eqref{eq:policy-optimization}. Then $\pi_{\psi_{k+1}}\longleftarrow\pi_\psi$.
}
\label{alg:ps-airl}
\end{algorithm}

As illustrated in Section \ref{sec:problem-definition}, we formulate the traffic simulation as multi-agent system problem by taking every vehicle in the traffic system as an agent interacting with the environment and each other. Inspired by parameter sharing trust region policy optimization (PS-TRPO)~\cite{gupta2017cooperative}, we incorporate its decentralized parameter sharing training protocol with AIRL in Section \ref{sec:method-airl}, and propose the parameter sharing AIRL (PS-AIRL) that learns policies capable of simultaneously controlling multiple vehicles in complex traffic environment. In our formulation, the control is decentralized while the learning is not. Accordingly, we make some simple assumptions for the decentralized parameter sharing agents. See Appendix for details\footnote{The appendix will be released on the authors' website and arxiv.}.

Under the decentralized parameter sharing training protocol as in \cite{gupta2017cooperative}, our proposed PS-AIRL can be highly sample-efficient since it reduces the number of parameters by a factor $M$, and shares experience across all agents in the environment. We use TRPO~\cite{Schulman2015Trust} as our policy optimizer, which allows precise control of the expected policy improvement during the optimization. For a policy $\pi_\psi$, at each iteration $k$, we perform an update to the policy parameters $\psi$ by solving the following problem:
\begin{eqnarray}
\label{eq:policy-optimization}
\begin{split}
    \min_\psi \ \  \mathbb{E}_{s,a\sim\pi_{\psi_{k}}, m\in\mathcal{M}} &\left[\frac{\pi_{\psi}(a | s, m)}{\pi_{\psi_{k}}(a | s, m)} A_{\psi_{k}}(s, m, a)\right]\\
    s.t., \ \  \mathbb{E}_{s \sim \pi_{\psi_{k}}} &\left[D_{\mathrm{KL}}\left(\pi_{\psi_{k}} \| \pi_{\psi}\right)\right] \leq \delta
\end{split}
\end{eqnarray}
where $\pi_{\psi_k}$ is the policy obtained in the previous iteration, $A_{\psi_k}(s,m, a)$ is an advantage function that can be estimated by the difference between the empirically predicted value of action and the baseline value. The training procedure of PS-AIRL is shown in Algorithm 1. The training pipeline details can be found in the appendix.

\nop{Fig. \ref{fig:framework} shows an overview of our dynamics-robust traffic simulation framework. Given a set of initial expert demonstrations, PS-AIRL learns an optimal policy and the reward function. During simulation, the learned policy plays the role of controlling the vehicles' behaviors. In the meantime, a TRPO step is applied to optimize the policy regularly every once in a while, so that the continuously updated policy can better fit the naturally non-stationary traffic environment. During this process, no expert demonstrations are needed any more.}

%% file: LearnSim/experiment.tex
\section{Experiment}
\label{sec:experiment}

\begin{table*}[t]

\fontsize{8.5pt}{8.5pt}\selectfont
\begin{center}
\begin{tabular}{C{2cm}|C{1cm}C{1cm}|C{1cm}C{1cm}|C{1cm}C{1cm}|C{1cm}C{1cm}|C{1cm}C{1cm}}
\toprule
\multirow{2}{*}{Method} & \multicolumn{2}{c|}{ HZ-1} & \multicolumn{2}{c|}{HZ-2} & \multicolumn{2}{c|}{HZ-3} & \multicolumn{2}{c|}{GD}& \multicolumn{2}{c}{LA } \\
 & Pos & Speed & Pos & Speed & Pos & Speed & Pos & Speed & Pos & Speed \\
 \specialrule{0.01em}{1.5pt}{2pt}
 CFM-RS & 173.0	&5.3	&153.0	&5.6	&129.0	&5.7	&286.0	&5.3 & 1280.9 & 10.3\\ 
 CFM-TS & 188.3	&5.8	&147.0	&6.1	&149.0	&6.1	&310.0	&5.5 & 1294.7 & 10.8\\ 
 \ours & \textbf{120.9}	&\textbf{4.5}	&\textbf{41.0}	&\textbf{2.1}	&\textbf{10.7}	&\textbf{1.1}	&\textbf{45.6}	&\textbf{1.0} &\textbf{681.5} & \textbf{5.1}\\ \specialrule{0.01em}{1.5pt}{2.0pt}
Improvement & 30.1\% & 15.1\% & 72.1\% & 47.6\% & 91.7\% & 80.7\% & 84.1\% & 81.1\% & 46.8\% & 50.5\% \\
\bottomrule
\end{tabular}
\caption{Performance comparison of \ours with state-of-art methods in terms of RMSE of recovered vehicle position (m) and speed (m/s).}
\label{tab:Exp1}
\end{center}
\end{table*}


We generally aim to answer the following three questions. 
\begin{itemize}
\vspace{-0.3em}
    \item \textbf{Q1}: \textit{Trajectory simulation}. Is \ours better at imitating the vehicle movement in expert trajectories?
    \vspace{-0.3em}
    \item \textbf{Q2}: \textit{Reward recovery}. Can \ours learn the vehicle's objective by recovering the reward function?
    \vspace{-0.3em}
    \item \textbf{Q3}: \textit{Capability to generalize}. Is the recovered reward function robust to dynamics changes in the traffic environment?
\end{itemize}

\subsection{Data and Experiment Setting}
We evaluate our proposed model on 3 different real-world traffic trajectory datasets with distinct road network structures collected from Hangzhou of China, and Los Angeles of US, including 3 typical 4-way intersections, a $4\times 4$ network, and a $1\times 4$ arterial network. See Appendix for details.

We compared \ours to two traditional CFM-based methods, CFM-RS and CRF-TS~\cite{krauss1998microscopic}, and three imitation learning-based models, BC~\cite{michie1990cognitive}, DeepIRL~\cite{wulfmeier2015maximum} and MA-GAIL~\cite{zheng2020learning}. See Appendix for details.

We use root mean square error (RMSE) to evaluate the mismatch between ground truth traffic trajectories and the trajectories generated by the simulation model. In addition, to verify the capacity of \ours in recovering the underlying reward that can generalize to different dynamics, we compare the reward values each model reaches given a hand-crafted reward function. See Appendix for its definition.

\nop{
\begin{figure*}[t]
\centering
 \includegraphics[width=1\textwidth]{fig/Fig_performance.pdf} 
\vspace{-1.5em}
\caption{RMSE of position and speed v.s. simulation time. RMSE increase w.r.t increasing simulation time for all methods, but \ours stays the best.}
\label{fig:performance}
\end{figure*}
}

%% file: LearnSim/discussion.tex
\begin{figure*}[t]
\centering
\footnotesize
 \includegraphics[width=0.94\textwidth]{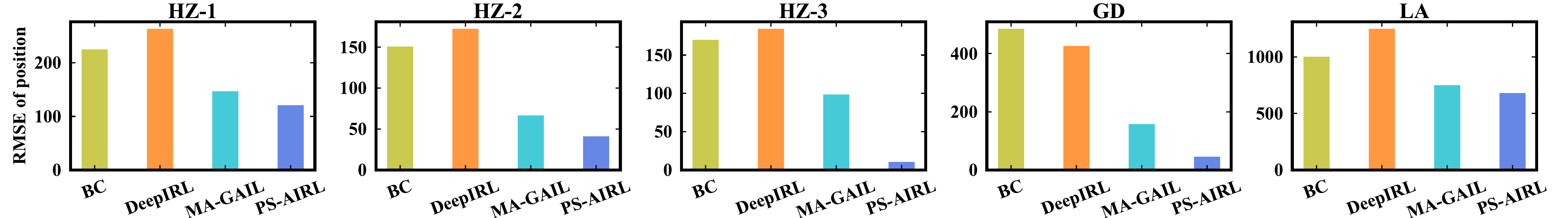} 
 \includegraphics[width=0.94\textwidth]{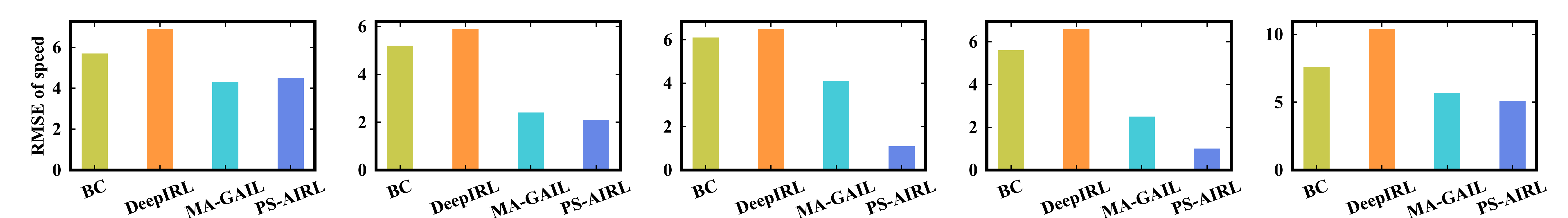} 
\caption{Performance comparison of variants of learning methods in terms of RMSE of position. \ours beats all alternatives. }
\label{fig:exp1-learning}
\end{figure*}

\begin{figure*}[t]
\centering
 \includegraphics[width=0.94\textwidth]{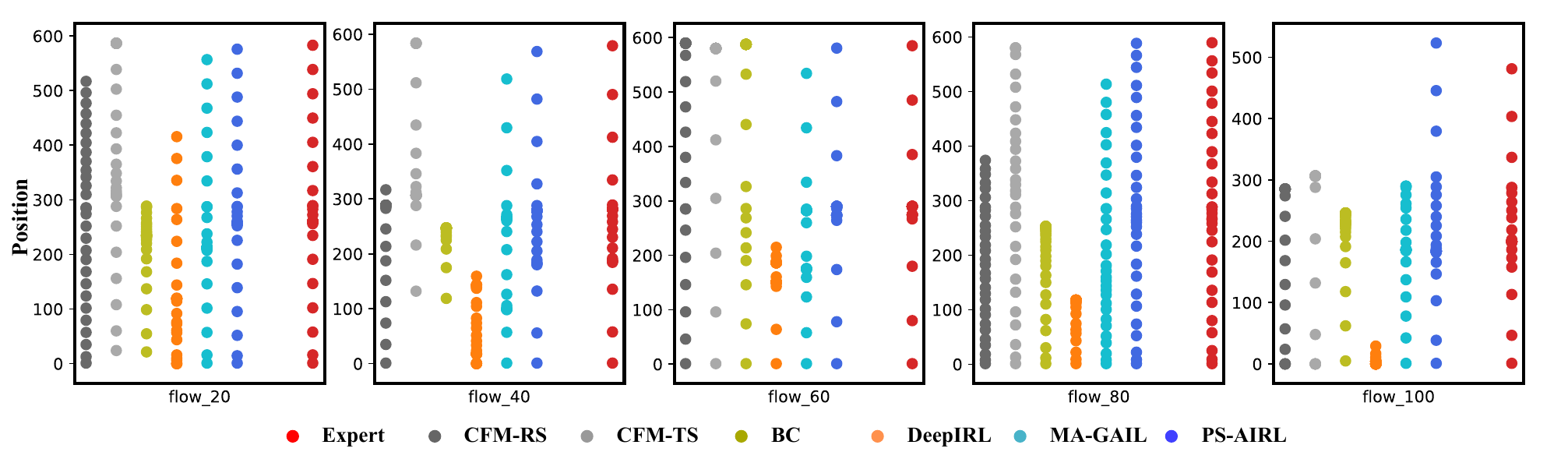} 
\caption{Trajectories of vehicles recovered by different methods on \hz-3. We arbitrarily select the vehicles with id 20, 40, 60, 80 and 100 and show their position w.r.t time. The position of vehicles recovered by \ours (blue) is most similar to the Expert (red).}
\label{fig:position-time}
\end{figure*}

\subsection{Traffic Simulation}
\label{sec:exp-traj-sim}
To answer Q1, we evaluate each model by using them to recover traffic trajectories, as shown in Table~\ref{tab:Exp1} and Figure~\ref{fig:exp1-learning}.

We first compare our proposed algorithm with {state-of-art simulation algorithms} \cfmrs and \cfmts (as in Table \ref{tab:Exp1}). We can observe that our proposed \ours outperforms the baseline methods on all datasets. \nop{Fig.~\ref{fig:performance} shows the performance of each method with respect to different lengths of simulation time. In most datasets, generally the trajectory mismatch tends to increase with time, due to the accumulation of action error. This effect is more obvious on RMSE of position. But it is evident that \ours shows a persistent and relatively stable good performance across all datasets.}
The two traditional methods \cfmrs and \cfmts  perform poorly due to the generally inferior fitting ability of physical CFM models against sophisticated traffic environment. We further compare \ours with state-of-art {imitation learning} methods. The results are shown in Figure \ref{fig:exp1-learning}. We can observe that \ours performs consistently better than other baselines. Specifically, as expected,  \ours and \gail perform better than the other two, for they can not only model the sequential decision process (compared to \bc) and utilize the adversarial process to improve the imitation (compared to \deepirl). More importantly, \ours achieves superior performance over \gail (except in \hz-1 these two perform similarly in terms of RMSE of speed). It indicates that our proposed model benefits from the learned invariant reward. For instance, if one vehicle enter the system right after another, we may not expect it to accelerate a lot. In contrast, if one vehicle enter the system with no preceding vehicle, it may speed up freely within the speed limit. Behaviors may seem different under these two cases, but the reward function still remains the same. Therefore, armed with explicit reward learning, \ours is able to outperform.

\textbf{Simulation Case Study} To have a closer look at the simulation of vehicle behavior, we extract the learned trajectory of 5 vehicles (vehicles with id 20, 40, 60, 80, 100) from \hz-3 as examples. Comparison between trajectories generated by different methods is shown in Fig. \ref{fig:position-time}. The trajectories recovered by \ours are most similar to the expert ones.

\subsection{Objective Recovery}
\label{sec:exp-reward-recovery}

To answer Q2, we leverage a hand-designed reward function as the ground truth and let the algorithms recover it. The reward increases when the vehicle approaches its desired speed and succeed in keeping safe distance from its preceding vehicles (see Appendix for details). Specifically, we use this hand-designed reward function to train an expert policy and use the generated expert trajectories to train each method. Note that, CFM-based methods are not included here for the following reasons. CFM-based methods can yield very unreal parameters (e.g., acceleration of 10 $m/s^2$) to match the generated trajectory. These will cause collision of vehicles or even one vehicle directly pass through another. This is an unreal setting. Hence, these CFM-based methods can not recover the true policy that vehicles follow in the simulation.

\textbf{Achieved Reward Value} We conduct experiments on \hz-1, \hz-2, \hz-3 and \gd, and the results are shown in Table \ref{tab:Exp2}. As expected, \ours achieves higher reward than other baselines and approach much closer to the upper bound reward value provided by the Expert. This indicates the better imitation (i.e. simulation) power of \ours.

\begin{table}[t]

\small
\begin{center}
\footnotesize
\begin{tabular}{C{1.4cm}C{1.1cm}C{1.1cm}C{1.1cm}C{1.1cm}}
\toprule
Method & HZ-1 & HZ-2 & HZ-3 & GD \\
 \specialrule{0.01em}{1.5pt}{1.5pt}
 \bc & -0.345 & -0.277 & -0.248 & -0.349\\
 \deepirl & -0.586 & -0.238 & -0.311  & -0.360\\
 \gail & -0.524 & -0.210 & -0.228 & -0.181\\
 \ours & \textbf{-0.173} & \textbf{-0.203} & \textbf{-0.201} & \textbf{-0.161}\\\specialrule{0.01em}{1.0pt}{1.0pt}
 Expert  & -0.060 & -0.052 & -0.062 & -0.065\\
\bottomrule
\end{tabular}
\caption{Achieved reward of different imitation learning methods. Higher reward indicates better imitation of the expert.}
\label{tab:Exp2}
\end{center}
\end{table}

\begin{figure}[t]
\centering
 \includegraphics[width=0.99\linewidth]{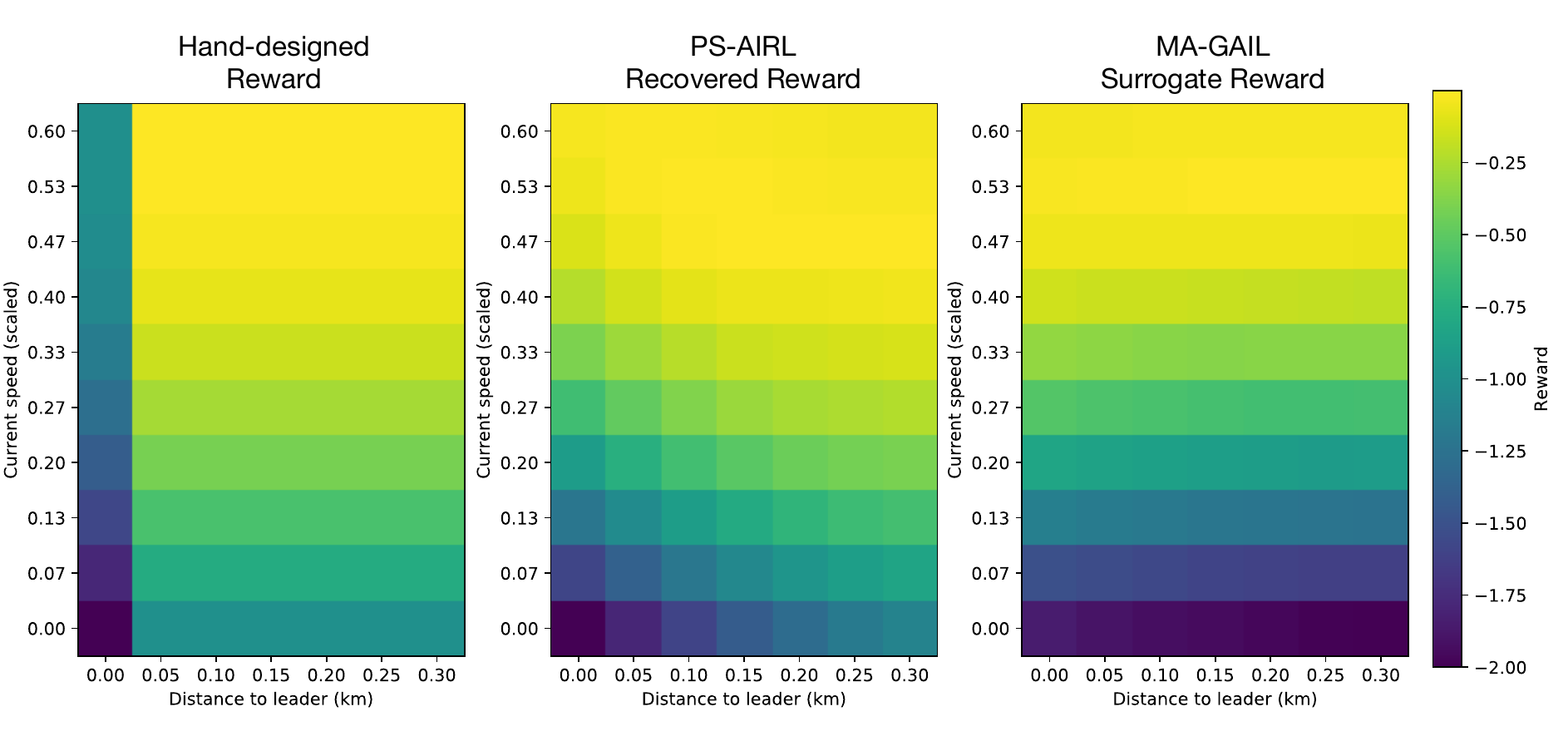}
\caption{Hand-designed reward (groundtruth), recovered reward of \ours and surrogate reward (output of discriminator) of \gail w.r.t. current speed and gap, on HZ-1. \ours can better infer the groundtruth reward function.}
\label{fig:recoverd-reward}
\end{figure}

\textbf{Recovered Reward Function} We are also interested in whether \ours can recover the groundtruth reward function. Hence, we visualize the reward function learned by \ours and the surrogate reward of \gail (the best baseline). Due to the space limit, results from other methods are omitted. They generally show even worse results than \gail. We enumerate the reward values for different speed and gap values. (Other state feature values are set to default values.) 
As in Fig.~\ref{fig:recoverd-reward}, compared to \gail, \ours recovers a reward closer to the groundtruth yet with a smoother shape. Our designed reward function applies penalties on slow speed and unsafe small gap. So the expert policy trained with this groundtruth reward tends to drive the vehicle on a faster speed and keep it against the front vehicle out of the safety gap. It shows that \ours is able to learn this objective of the expert in choosing optimal policies from demonstrations, assigning higher reward values to states with faster speed and safe gap, while applying penalties on the opposite. In contrast, the surrogate reward learned by the \gail totally ignores the safe gap. In addition, when vehicles are out of the safe gap, \ours recovered a reward value closer to the groudtruth. 

\subsection{Robustness to Dynamics}
\label{sec:exp-generalize}

To answer Q3,  we transfer the model learned in Section~\ref{sec:exp-reward-recovery} to a new system dynamics. Specifically, compared with Section~\ref{sec:exp-reward-recovery}, we change the maximum acceleration and deceleration of vehicles from initially 2 $m/s^2$ and 4 $m/s^2$ into 5 $m/s^2$ and 5 $m/s^2$ respectively. Other settings remain unchanged. Because the reward function remains unchanged (i.e., drivers still want to drive as fast as possible but keep a safe gap), reward-learning methods should be able to generalize well. 

For \deepirl, \gail and \ours, we use the learned reward function (surrogate reward function for \gail) to train a new policy. (\bc policy is directly transferred because it do not learn reward function.) We also include a direct transfer version of \ours as comparison to demonstrate the necessity of re-training policy under new dynamics. Similar to the reward recovery experiment, CFM models also do not apply to this comparison.

\begin{table}[t]

\small

\begin{center}
\begin{tabular}{C{2cm}C{1cm}C{1cm}C{1cm}C{1cm}}
\toprule
Method & HZ-1 & HZ-2 & HZ-3 & GD \\
 \specialrule{0.01em}{1.5pt}{1.5pt}
 \bc ($\mathcal{D}$) & -0.406 & -0.322 & -0.336 & -0.681\\
 \deepirl &  -0.586 & -0.288 & -0.362 & -0.369\\
 \gail &  -0.483 & -0.243 & -0.337 & -0.175 \\
  \ours ($\mathcal{D}$) & -0.448 & -0.254 & -0.321 & -0.598 \\
 \ours &  \textbf{-0.384} & \textbf{-0.192} & \textbf{-0.248} & \textbf{-0.161}\\ 
\bottomrule
\end{tabular}
\caption{Reward under new dynamics. \bc ($\mathcal{D}$) and \ours ($\mathcal{D}$) means directly applying the learned policy. The other methods use the learned reward to retrain a policy under new dynamics.}
\label{tab:Exp3}
\end{center}
\end{table}

Table \ref{tab:Exp3} shows the achieved reward in the new environment. 
\ours achieves highest reward, because with the invariant reward function learned from the training environment, \ours can learn a new policy that adapts to the changed environment. \deepirl perform poorly in transferred environment because of its incapability of dealing with large number of states. \ours ($\mathcal{D}$) fails to generalize to the new environment, which indicates that the policy learned in the old dynamics are not applicable anymore. \gail, though retrained with the surrogate reward function, can not transfer well, due to the dependency on the system dynamics.

%% file: LearnSim/conclusion.tex
\section{Conclusion}
\label{sec:conclusion}

In this paper, we formulated the traffic simulation problem as a multi-agent inverse reinforcement learning problem, and proposed \ours that directly learns the policy and reward function from demonstrations. Different from traditional methods, \ours does not need any prior knowledge in advance. It infers the vehicle's true objective and a new policy under new traffic dynamics, which enables us to build a dynamics-robust traffic simulation framework with PS-AIRL. Extensive experiments on both synthetic and real-world datasets show the superior performances of PS-AIRL on imitation learning tasks over the baselines and its better generalization ability to variant system dynamics.

%% file: LearnSim/acknowledment.tex
\section*{Acknowledgment}
This work was supported in part by NSF awards \#1652525, \#1618448 and \#1639150. The views and conclusions contained in this paper are
those of the authors and should not be interpreted as representing
any funding agencies.

%% file: LearnSim/appendix.tex
\begin{appendices}
\appendixpage

\section{Simulation Environment}

We created an OpenAI Gym\footnote{https://gym.openai.com/} environment as our simulation environment based on the a traffic simulator called CityFlow\footnote{https://cityflow-project.github.io}~\cite{zhang2019cityflow}. 
CityFlow is an efficient traditional car-following model (CFM) based traffic simulator that is able to conduct 72 steps of simulations per second for thousands of vehicles on large scale road networks using 8 threads. The CFM used in CityFlow is a modification of the Krauss car-following model with the key idea that, the vehicle will drive as fast as possible subject to perfect safety regularization. The vehicles in CityFlow are subject to several speed constraints, and will drive in the maximum speed that simultaneously meets all these constraints.

This CityFlow-based environment is able to provide multiple control APIs that allow us to obtain observations and give vehicles direct controls that bypasses the embedded default CFM. At each timestep, for a certain vehicle, a state representation is generated from the environment. The policy will determine the best action (i.e., next step speed) according to the state and conduct it on the environment. Since we do not have trajectory data in granularity of seconds, on the HZ and GD roadnets in our experiments (Section VI), we use CityFlow to generate trajectories as demonstration data based on real traffic flow. In these demonstration datasets, the embedded CFM in CityFlow serves as an expert policy. But during simulation learning, the embedded CFM has no effect on the learned policy. For the LA dataset, we directly extract the vehicle trajectories from the dataset.

This environment is initiated with a road network, traffic flow, and a signal plan. The road network includes information of intersections, roads, and length of each lane. The traffic flow provides the track of each vehicle, as well as the time and location it enters the system. The traffic signal plan gives the phase of traffic light for each intersection.

\section{Dataset}
\label{subsec:dataset}

\textbf{Hangzhou (HZ)}. This dataset covers three typical 4-way intersections in Hangzhou, China, collected from surveillance cameras near intersections in 2016. Each record contains the vehicle id, arriving time and location. Based on this, we use the default simulation environment of CityFlow~\cite{zhang2019cityflow} (in Section 5.2) or trained by a hand-designed reward function (in Section 5.3 and 5.4) to generate expert demonstration trajectories, denoted to as HZ-1, HZ-2, and HZ-3.

\begin{figure}[t]
 \includegraphics[width=0.49\textwidth]{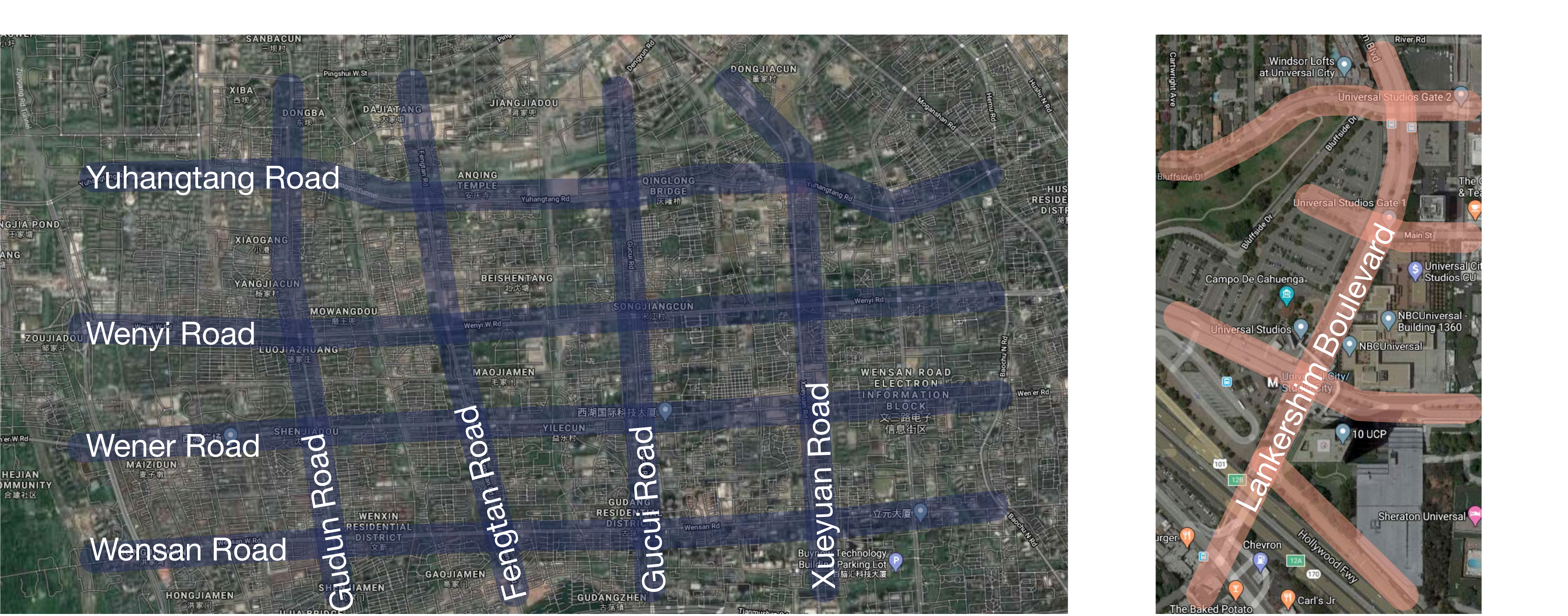}
  \begin{tabular}{p{18em}c}
   \hspace{13mm} (a) Gudang, Hangzhou  & (b) LA \\
 \end{tabular}
 \caption{Maps of the $4\times 4$ roadnet of Gudang Area, Hangzhou and the $1\times 4$ roadnet of Lankershim Boulevard, LA used in our experiment.}
\label{fig:roadnet}
 \end{figure}

\textbf{Gudang (GD)}. This a $4 \times 4$ network of Gudang area (Figure \ref{fig:roadnet} (a)) in Hangzhou, collected in same fashion as \hz dataset. This region has relatively dense surveillance cameras. Necessary missing data fill in are conducted. We process it in the same way as \hz dataset. 

\textbf{Los Angeles (LA)}. This is a public traffic dataset~\footnote{https://ops.fhwa.dot.gov/trafficanalysistools/ngsim.htm} collected from Lankershim Boulevard, Los Angeles on June 16, 2005. It covers an $1 \times 4$ arterial with four successive intersections (shown in Figure~\ref{fig:roadnet} (b)). This dataset records the position and speed of every vehicle at every 0.1 second. We directly use these records as expert trajectories.

\section{Compaired Baselines}

\textbf{Traditional CRF-based Method}
\begin{itemize}
    \item \textbf{CFM-RS}: the CFM calibrated with random search. In each trial, a set of parameters for the Krauss model is randomly selected given a limited searching parameter space. After limited times of trials, the set of parameters that yields trajectories most similar to demonstrations is kept as the calibration result.
    \item \textbf{CFM-TS}: the CFM calibrated with Tabu search~\cite{osorio2019efficient}. Given an initial set of parameters, a new set is chosen within the neighbors of the current parameter set. The new set of parameters will be included in the Tabu list, if they generates better trajectories. The best set of parameters in the Tabu list is shown in the results.
\end{itemize}
\textbf{Imitation Learning Method}
\begin{itemize}
    \item \textbf{BC}: behavioral cloning~\cite{michie1990cognitive} is a typical imitation learning method with a supervised learning procedure. It directly trains a policy network to fit the state-action mapping. 
    \item \textbf{\deepirl}: maximum entropy deep IRL~\cite{wulfmeier2015maximum}. Since it is unable to deal with large state space, in our experiments, we use spectral clustering~\cite{ng2002spectral} to discretize the state extracted from the environment into 64 states as the input of \deepirl.
    \item \textbf{MA-GAIL}: An extension of GAIL~\cite{zheng2020learning} for traffic simulation. 
\end{itemize}
\nop{For all the learning methods, we use neural networks in similar sizes and structures to represent each component, in order for fair comparison.}
\nop{For BC and \deepirl, with no control during training, we simply use all trajectories to train a single policy network, regardless of the index.}

\begin{figure*}[t!]
\centering
 \includegraphics[width=0.9\textwidth]{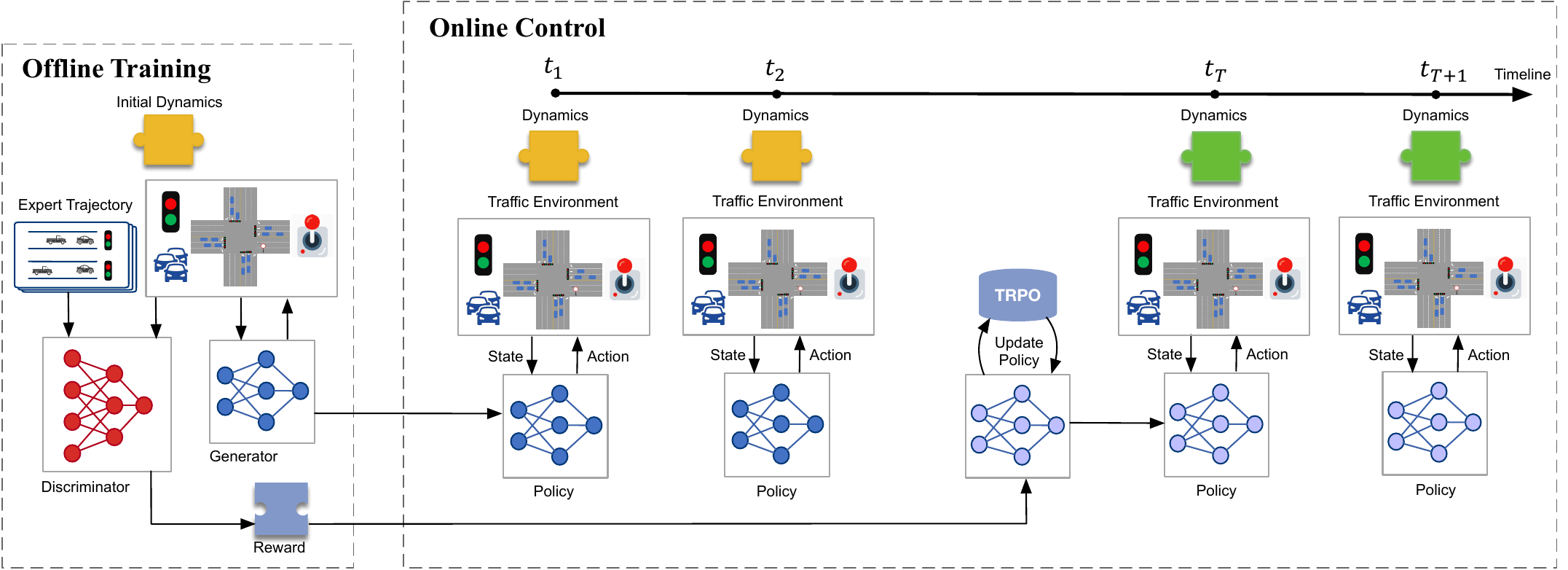}
\caption{The dynamics-robust framework of our PS-AIRL simulation model. The policy and reward are both learned during the offline training. During online control of the simulator, the learned policy plays the role of controlling the movement of vehicles in the environment. We employ TRPO to regularly optimize a new policy with the learned reward function. No expert demonstrations are needed for policy re-optimization.}
\label{fig:framework}
\end{figure*}

\section{Evaluation Metrics}
\label{subsec:metric}
We evaluate the trajectories generated by the policies of each method against the expert trajectories, using root mean square error (RMSE) of position and speed. The RMSE of position is defined as $RMSE = \frac{1}{T}\sum_{t=1}^{T}\sqrt{\frac{1}{m}\sum_{j=1}^{m}\left(s_j^{(t)} - \hat{s}_j^{(t)}\right)^2}$,
where $T$ is total time of the simulation, $s_j^{(t)}$ and $\hat{s}_j^{(t)}$ are the true value and observed value of the position of the $j$-th vehicle at timestep $t$, respectively. The RMSE of speed is calculated similarly.

\nop{\begin{equation}
\begin{split}
RMSE_{pos} = \sqrt{\frac{1}{m}\sum_{j=1}^{m}\left(s_j - \hat{s}_j\right)^2}\\
RMSE_{speed} = \sqrt{\frac{1}{m}\sum_{j=1}^{m}\left(v_j - \hat{v}_j\right)^2}
\end{split}
\end{equation}
where $s_j$ and $v_j$ are the true values of the position and speed of the $j$-th vehicle at current timestep, and $\hat{s}_j$ and $\hat{v}_j$ are the simulated values of the position and speed of the $j$-th vehicle at the current timestep. We further calculate the average of RMSE over time as the final measurement.}


\subsubsection{Hand-crafted Reward}

In order to verify the capability of \ours to recover reward function and generalize to different dynamics, we further compare the reward values that different methods can reach. We use the hand-crafted reward function defined in Eq.~\ref{eq:reward-function} as the ground truth reward. For each vehicle at each timestep, we define its reward $r$ as the opposite of the match error of its speed $v$ and gap $g$ (distance to the preceding vehicle or the traffic light) against the desired speed $v_{des}$ and gap $g_{des}$.
\begin{equation}
    r = -\left[\left(\frac{v - v_{des}}{v_{max}}\right)^2 + \lambda \left(\frac{g - g_{des}}{g_{min}}\right)^2 \right]
    \label{eq:reward-function}
\end{equation}
where $\lambda$ is a trade-off parameter, $v_{max}$ and $g_{min}$ are the speed limit in the current lane and the given minimum gap between vehicles. The $v_{des}$ and $g_{des}$ are determined according to the traffic signal status, speed limit and distance to the leading vehicle.  In Section 5.3 and 5.4, we use the mean reward value $\Bar{r}$ (first averaged by a vehicle's episode length, and then averaged by the total number of vehicles in system) as the evaluation metric.

\begin{figure}[t]
\centering
 \includegraphics[width=0.4\textwidth]{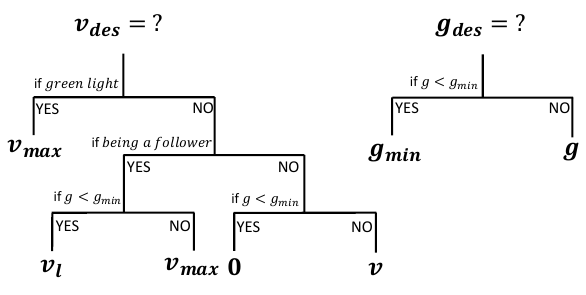} 
\caption{Calculation of desired speed $v_{des}$ and desired gap $g_{des}$ in the hand-designed reward function, where $v$ and $v_l$ are the velocity of the focused vehicle and its preceding vehicle, $v_{max}$ and $g_{min}$ is the given speed limit and desired minimum gap, and $g$ is the gap against the front car.}
\label{fig:reward}
\end{figure}

For the hand-crafted reward, we determine the desired speed and the desired gap as in Figure~\ref{fig:reward}. Basically, the vehicle desires to keep a speed as high as the max speed when possible, and to decelerate to a safe speed if it is too close to a preceding vehicle or a red traffic signal. In addition, the the vehicle aims to maintain a gap from the preceding vehicle larger or equal to the minimum gap. 
In Section 5.3, we set $\lambda\!=\!1$, $v_{max}=11 m/s$ and $g_{min}=25 m$.

\begin{table}[t]
\centering
\small
\begin{tabular}{L{2.5cm}L{3.5cm}}
\toprule
\textbf{Feature Type} & \textbf{Detail Features}\\
\midrule
\multirow{1}{*}{Road network}  & Lane ID, Length of current Lane, Speed limit\\
\specialrule{0.01em}{1pt}{1pt}
Traffic signal & Phase of traffic light\\
\specialrule{0.01em}{1pt}{1pt}
\multirow{1}{*}{Vehicle} & Velocity, Position in lane, Distance to traffic light\\
\specialrule{0.01em}{1pt}{1pt}
\multirow{1}{*}{Preceding vehicle} & Velocity, Position in lane, Gap to preceding vehicle\\
\specialrule{0.01em}{1pt}{1pt}
\multirow{1}{*}{Indicators} & If leading in current lane, If exit from intersection\\
\bottomrule
\end{tabular}
\caption{Observation features.}
\label{tab:features}
\end{table}

\section{Reward and System Dynamics}
\begin{figure}[t]
\centering
 \includegraphics[width=0.6\linewidth]{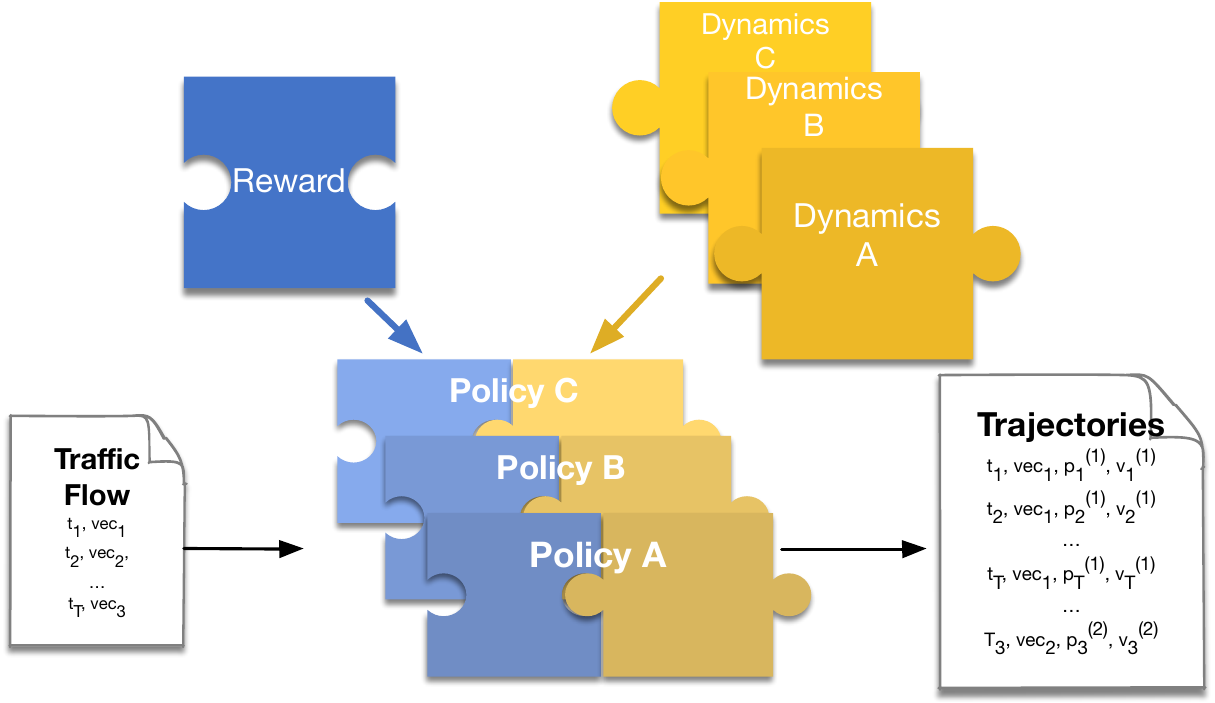} 
\caption{Reward, system dynamics and policy. Ideally, the reward with distinct system dynamics remains the same.}
\label{fig:reward-dynamics-policy}
\end{figure}

Similar to human driver's decision making, reward function and system dynamics are the two key components that determines a vehicle's driving policy (as illustrated in Figure~\ref{fig:reward-dynamics-policy}).
The reward function can be view as how the a vehicle values different state components during driving, e.g., safety and efficiency. For instance, a real-world vehicle usually attempts to drive as fast as it can under the assumption that it is safe (e.g., keeping a safe distance to the preceding vehicle). A driving policy is formed by selecting actions that maximize their expected reward.
System dynamics, on the other hand, describes how the vehicle's actions yield the next state of environment. System dynamics consist of some physical factors including road conditions (e.g., speed limit, road surface and weather condition) and vehicle properties (e.g., maximum speed, maximum acceleration and maximum deceleration). The difference of these factors in different time and locations may lead to the change of system dynamics.

\section{Assumptions}
In our formulation, the control is decentralized while the learning is not~\cite{gupta2017cooperative}. Accordingly, we make simple assumptions for our decentralized parameter sharing agents as follows.
\begin{itemize}
    \item \textbf{Agent}: Each vehicle as an independent agent $m\in\mathcal{M}$, cannot explicitly communicate with each other but can learn cooperative behaviors only from observations.
    
    \item \textbf{State} and \textbf{Action}: We select a set of features to define observations (partially observed state), as shown in Table \ref{tab:features} in Appendix. We use speed as action for each vehicle agent. In each timestep the agent makes a decision in choosing a speed for the next timestep. Each agent has the same state and action space:
        $$\mathcal{S}_m\in\mathbb{R}^S \ \text{and}\  \mathcal{A}_m\in\mathbb{R}^A, \forall m\in\mathcal{M}$$
    
    \item \textbf{Policy}: All agents share the parameters of a homogeneous policy $\pi(a|s)$. This enables the policy to be trained with the experiences of all agents at the same time, while it still allows different actions between agents since each agent has independent different observations.

    \item \textbf{Reward}: All agents share the parameters of a homogeneous unknown reward function $r(s,a)$. Accordingly, each vehicle agent gets distinct reward given different observations and actions.
\end{itemize}

\section{Network Architectures}

Our whole model can be divided into two parts, discriminator and generator. The network structure and size are shown in Table~\ref{tab:network-detail-para}.

\textbf{Discriminator.} In our \ours, the discriminator is composed of two networks, the reward function $\hat{r}_\theta$ and the shaping term $h_\phi$, as well as a sigmoid function. Concretely, the reward function $\hat{r}_\theta$ only takes state as input and output the estimation of the disentangled reward for the current state. The shaping term $h_\phi$ also takes state features as input, as a regularization to the reward $\hat{r}_\theta$ that helps mitigate the reward ambiguity effect. The whole network of the discriminator $D_{\theta,\phi}$ can be presented as follows

\begin{equation}
\label{eq:learned-func-appendix}
\begin{split}
\hat{r}_\theta(s) & = \text{ReLU-Net}_\theta(s)\\
h_\phi(s) & = \text{ReLU-Net}_\phi(s)\\
f_{\theta,\phi}(s,a,s') & =\hat{r}_\theta(s) + \gamma h_\phi(s')-h_\phi(s)\\
D_{\theta,\phi}(s,a,s') & = \sigma\left(f_{\theta,\phi}(s,a,s') - \log\pi(a|s)\right)
\end{split}
\end{equation}
where $s$ is the current state, $s'$ is the state for the next timestep, $\text{ReLU-Net}(\cdot)$ is a ReLU network with two hidden layers, 
$\sigma(\cdot)$ is a sigmoid function, and
$\pi$ is the learned policy during the last iteration.

\textbf{Generator (policy).} The generator is using an actor-critic agent optimized by trust region policy optimization (TRPO). In detail, we use Beta distribution in the actor network, as it can restrict the range of the output action. The actor network takes state as input, and outputs the values of two parameters for the distribution. Then we sample action from the distribution determined by these two parameters. For the critic network, it takes the state features as input and outputs the estimation of the value of current state.

Some of the key hyperparameters are shown in Table~\ref{tab:hyper}.

\begin{table}[h]
    \centering
    \caption{Summary of network structures. FC stands for fully-connected layers. Here, ($11+M$,) is the dimension for the state features, where $M$ is the number of lanes in the environment. This is because we use one-hot encoding for the lane id feature. }
    \fontsize{8.0pt}{8.0pt}\selectfont
    \begin{tabular}{lllll}
    \toprule
    \textbf{Network}     & $\hat{r}_\theta$ & $h_\phi$ & Actor & Critic \\
    \midrule
    Input & ($11+M$,)  & ($11+M$,)  & ($11+M$,)  & ($11+M$,) \\
    \midrule
    \multirow{2}{*}{Hidden layers} & FC(32) & FC(32) &  FC(32) & FC(32) \\
     & FC(32) & FC(32) & FC(32) & FC(32) \\
    \midrule
    Activation & ReLU & ReLU & ReLU & ReLU\\
    \midrule
    Output & (1,) & (1,) & (2,) &  (1,) \\
    \bottomrule
    \end{tabular}
    \label{tab:network-detail-para}
\end{table}

\begin{table}[h]
    \centering
    \caption{Summary of hyperparameters}
    \begin{tabular}{ll}
    \toprule
    \textbf{Hyperparameters}     & \textbf{Value}  \\
    \midrule
    Update period (timestep) & 50 \\
    Discriminator update epoch & 5 \\
    Generator (policy) update epoch & 10 \\
    Batch size & 64\\
    Learning rate $\alpha$ & 0.001\\
    \bottomrule
    \end{tabular}
    \label{tab:hyper}
\end{table}

\section{Details of Training Pipeline}

Fig. \ref{fig:framework} shows an overview of our dynamics-robust traffic simulation framework. Given a set of initial expert demonstrations, PS-AIRL learns an optimal policy and the reward function. During simulation, the learned policy plays the role of controlling the vehicles' behaviors. In the meantime, a TRPO step is applied to optimize the policy regularly every once in a while, so that the continuously updated policy can better fit the naturally non-stationary traffic environment. During this process, no expert demonstrations are needed any more.

\end{appendices}